\documentclass[runningheads]{llncs}

\usepackage[T1]{fontenc}
\usepackage{graphicx}
\usepackage{booktabs}
\usepackage{amsmath}
\usepackage{float}
\usepackage{tikz}
\usetikzlibrary{arrows.meta,positioning}
\usepackage[style=numeric,backend=biber]{biblatex}
\graphicspath{{figures/}}

\title{How Well Does Webcam Eye Tracking Work ``in the Wild''? A Large-Scale Validation of Chiasm and a Comparison with WebGazer}
\titlerunning{Webcam Eye Tracking ``in the Wild''}

\author{Artem V. Belopolsky, Florian Perdreau}
\authorrunning{Belopolsky and Perdreau}
\institute{Chiasm, BV, Amsterdam, The Netherlands}

\begin{document}
\maketitle
\begin{abstract}
Webcam eye tracking can greatly accelerate data collection in basic and applied research by reaching large, diverse samples of participants online.
However, its spatial performance is generally considered well below that of hardware-based systems, limiting its use. We evaluated the 
first-generation Chiasm webcam eye tracker in a large online study ($N = 505$) spanning diverse participants, devices, and recording conditions. 
Chiasm achieved a mean spatial accuracy of 2.39~cm ($\approx 2.28^{\circ}$ at 60~cm),
spatial precision (SD) of 1.05~cm ($\approx 1.00^{\circ}$ at 60~cm), and sample-to-sample precision (RMS-S2S) of 1.29~cm ($\approx 1.23^{\circ}$ at 60~cm). For the 91\% of participants who met a 3~cm calibration criterion,
spatial accuracy was 2.15~cm ($\approx 2.05^{\circ}$ at 60~cm). After three minutes of a task that encouraged head movement, spatial accuracy declined by 0.59~cm, while
the gaze center at each target shifted by a mean distance of only 0.37~cm. 
On the validation block, the offset was 0.30~cm larger for men than for women. Age, region of residence, face luminance, and Michelson contrast showed no reliable relationship with spatial accuracy. A higher eye aspect ratio, a more open eye, was associated with a smaller offset. Wearing glasses decreased spatial accuracy on every validation block. In a comparison between a random subsample ($N = 101$) and a separate sample tested with
the open-source tracker WebGazer ($N = 101$) using an identical protocol, Chiasm showed substantially better spatial accuracy (2.18 vs.\ 4.51~cm) and spatial precision (SD; 1.01 vs.\ 1.55~cm).
WebGazer showed better sample-to-sample precision (RMS-S2S; 0.79 vs.\ 1.26~cm), which likely reflects conservative, over-smoothed
estimates rather than superior stability. These results suggest that Chiasm is narrowing the gap between webcam and hardware-based eye tracking, making 
large-scale gaze research ``in the wild'' a reality.
\end{abstract}


\section{Introduction}
\label{sec:introduction}

Eye tracking has proven to be a powerful tool for studying human behavior and cognition, with applications ranging from basic research to user experience (UX)
research, marketing, and clinical practice \cite{yang2021webcam,slim2023moving,prystauka2024online,fang2026investigating,steffan2024validation,vos2022comparing,hutt2022evaluating,banki2022comparing,ozsoy2023turkish,vandercruyssen2024validation,slim2024webcams,afifah2025exploratory,james2025paradigms,bogdan2024investigating,bramlett2025wrangling,hutt2024webcam}. Yet its potential remains constrained by how difficult it is to collect large and diverse datasets.
Most eye tracking is still conducted in the lab, under controlled conditions, one participant at a time. Wearable eye trackers have relaxed some of these
constraints, but they still rely on specialized hardware that must be delivered to participants. Webcam eye tracking offers a way around this bottleneck by
using the commodity webcams built into virtually every laptop, tablet or smartphone. The idea is simple: anyone with such a device can take part in an eye-tracking experiment. Combined
with large, vetted online participant pools such as Prolific \cite{palan2018prolific}, this shrinks data collection from weeks to hours.

However, fulfilling this promise presents significant challenges, as data collection takes place ``in the wild'', where researchers cannot control participants’ 
hardware, environment, or behavior. First, commodity webcams are far less capable than the cameras used in lab-based systems. They have lower spatial 
resolution (often as low as $640 \times 480$ pixels, with the eye occupying only a small fraction of the image) and lower sampling rates (typically 30--60~Hz,
compared to 500--2000~Hz for lab systems such as the EyeLink 1000 Plus). They also lack infrared illumination, which rules out the pupil--corneal reflection
method on which most lab-based trackers rely. In addition, webcams apply automatic exposure, white balance, and compression that the experimenter cannot
control, and browsers introduce variable frame timing. The rest of the setup varies as well: computing hardware, operating systems, browsers, monitors, 
and internet connections all differ across participants. Second, the recording environment is uncontrolled: ambient lighting, viewing distance,
and potential distractions vary from one participant to the next and can change during a session. Finally, there is no control over participants' posture
and head movements, which are not constrained by a chin rest. Their engagement is also uncontrolled, and an experimenter cannot intervene in real time when
data quality deteriorates. These three major challenges for obtaining high-quality gaze data using webcam require not only a gaze estimation algorithm that
is robust to the variability in computing infrastructure, recording environment and participant behavior, but also a well-designed user interface that allows
self-guidance, feedback and error recovery for the participant, to compensate for the absence of an experimenter in the room.

The most widely used webcam eye tracker is WebGazer \cite{papoutsaki2016}, an open-source, JavaScript library that detects the eyes in the webcam 
image and maps eye-image features to screen coordinates using ridge regression, trained on clicks or gazing at calibration points. Several studies
estimated WebGazer's spatial accuracy to be around $4$--$5^{\circ}$ in the lab with further degradation online \cite{semmelmann2018}. Most commercial 
webcam eye
trackers, such as RealEye.io (based on WebGazer, \cite{wisiecka2022}) and LabVanced (deep learning-based, \cite{kaduk2024}), have not been 
validated ``in the wild'', but rather under controlled and favourable conditions against lab-based systems, and reported spatial accuracy of around $1.5^{\circ}$.
One recent study \cite{saxena2024} compared the performance of several deep learning-based models for gaze estimation (MPIIGaze \cite{zhang2015},
 ETHXGaze \cite{zhang2020} and FAZE \cite{park2019}) on webcam frames collected \textit{online} ({$N=65$}), but analyzed \textit{offline}, in several tasks. They showed
 that FAZE achieved the best performance, with a spatial accuracy of $2.44^{\circ}$ and a spatial precision (SD) of $1.63^{\circ}$ and RMS-S2S of $0.47^{\circ}$. 
 MPIIGAZe and ETHXGaze achieved a spatial accuracy of $3.70^{\circ}$ and $3.40^{\circ}$, a spatial precision (SD) of $2.96^{\circ}$ and $2.39^{\circ}$,
and RMS-S2S of $2.33^{\circ}$ and $1.80^{\circ}$ respectively. The increased performance for different models came at the expense of inference time, with 
inference latency of 12 fps for MPIIGaze, 4 fps for ETHXGaze and 1 fps for FAZE. Especially for FAZE, the most promising model, this means that all processing,
including the calibration procedure, has to be done offline, which may impair user experience and be impractical for some applications. It is worth noting
that this research evaluated the offline performance of several models on webcam data, but did not evaluate a webcam eye tracking system.

The largest validation of a webcam eye tracker ``in
the wild'' ($N=255$) has been recently conducted by iMotions with their own WebET 3.0 tracker \cite{imotions2023}. Their algorithm uses deep-learning model 
``pre-trained on thousands of faces'' and with a radial basis function (RBF) calibration mapping. They reported a spatial accuracy of $2.1^{\circ}$ and that the presence
of glasses was the only factor that significantly degraded spatial accuracy. However, the white paper report of iMotions does not provide enough details regarding the validation procedure and analysis, and does not report
metrics such as spatial precision (SD) or sample-to-sample precision (RMS-S2S), which are crucial for the evaluation of the tracker's performance and prevents 
a clear comparison to other trackers.

There are a few reasons to suspect that the metrics that were previously reported for the webcam eye tracking may be rather optimistic. First, 
most webcam eye trackers were tested under controlled laboratory conditions, which are not representative of diverse online scenarios
\cite{wisiecka2022}, \cite{kaduk2024}. 
Second, in some studies, recalibration was often performed after short intervals, which was done to reduce the effect of drift over time,
but it is not always practical when running online \cite{kaduk2024}, \cite{saxena2024}. Third, none of the studies estimated the drift in calibration metrics
over time, which is important to know and control for, as showed by \cite{ehinger2019} for the hardware-based eye trackers. Finally, many studies did not report 
all the metrics that are crucial for the evaluation of the 
webcam eye tracker's performance, such as spatial accuracy, spatial precision (SD) and sample-to-sample precision (RMS-S2S) \cite{semmelmann2018}, 
\cite{imotions2023}.

In this paper, we evaluated the first-generation Chiasm webcam eye tracker (\url{https://chiasm.eu}), which uses a deep neural network 
to predict gaze from a single webcam image, combined with a custom calibration procedure. We conducted a large online study ($N=505$) spanning diverse 
participants, devices, and recording conditions. We designed a novel validation protocol which allowed us to comprehensively evaluate the webcam
eye tracker's performance ``in the wild''. We report spatial accuracy, spatial precision (SD), sample-to-sample precision (RMS-S2S),
as well as drift of these metrics over time. In addition, 
we evaluated how the tracker's performance varies with different participant and environmental characteristics, such as age, gender, region of residence, 
eye aspect ratio, face luminance, or Michelson contrast, and presence of glasses. We also compared the performance of Chiasm with WebGazer, since the latter is the most
widely used webcam eye tracker and could be implemented with an identical experimental protocol. We collected a separate sample with WebGazer ($N = 101$) and 
compared it with a random sample of participants from the Chiasm cohort.

\section{Methodology}
\label{sec:methodology}
We used Chiasm's implementation in jsPsych \cite{deleeuw2023jspsych} and designed a novel experimental protocol that allowed us to compute various metrics of the webcam eye tracker's 
performance ``in the wild''. First, we gave 
participants three attempts to calibrate the tracker. This ensured that they had a chance to adjust to the calibration procedure and to give it their best
effort.
Second, even if they failed to calibrate the tracker after three attempts, we still allowed them to continue the session. This way we did not bias our
results by excluding participants who failed to calibrate the tracker. 
Third, directly after the initial 9-point calibration and validation, which used pulsating discs, designed to attract attention, we presented participants with
a 13-dot validation block (Validation A), with static discs, which are commonly used in lab-based eye tracking. 
Fourth, we presented participants with a shooting game, which encouraged them to move their head. Right after the shooting game, another 13-dot validation
block was presented (Validation B). This allowed us to assess the tracker's drift over time and after head movements. The whole experiment lasted
approximately 14 minutes.

\subsection{Participants}

The complete dataset (Chiasm and WebGazer) was gathered from multiple participant batches over 12 days, totaling 36 hours of active collection time. Participants were 
recruited via Prolific and gave informed consent. Data were processed under a data protection impact assessment (DPIA) by Chiasm B.V., in accordance with the General Data
Protection Regulation (GDPR). They took part remotely using a web browser. Participants using phones, tablets, iPadOS or Desktop Safari were not allowed to enter the
experiment.
Declining informed consent also ended the session, as did a failure to initialise the tracker. Participants who continued, reported whether the webcam was built-in, 
 external, or of unknown type; whether the computer was a laptop or a desktop; and whether the display was a laptop screen, an external monitor, or an all-in-one screen.

Appendix~\ref{app:chiasm-attrition} counts participants who started Validation~A, including those who did not meet the acceptance threshold, and how many of them started the shooting game, started Validation~B, and were included in the final sample.

\subsection{Apparatus}
\label{sec:apparatus}

Participants used their own webcam and display, without a chin rest, and viewing distance was left free. The sampling rate was capped at 30Hz and the frames were
rescaled to 640x480 pixels. Physical scale was obtained before calibration by matching an on-screen rectangle to a standard identification card, 85.6~mm wide. The match supplied a pixels-per-centimetre factor for the screen plane.
When the browser's page zoom was away from 100\%, a warning was shown and the participant could continue after resetting the zoom. The display was then placed in fullscreen. The tracker recorded webcam frames and the gaze prediction for each frame. The cursor was hidden during the validation blocks and the shooting game.

\subsection{Stimuli}
\label{sec:stimuli}

Calibration and its acceptance check used a 9-point grid on a white field: the screen center, the four edge midpoints, and the four corners. Peripheral points were inset 
by 10\% of screen width or height, and every point was kept at least 40~px inside the canvas. The center point appeared first; the other eight followed in random order.
 Each point was combination of bulls eye and cross hair, known to minimize microsaccades \cite{thaler2013}. Each point was a red disc that expanded from a radius
 of 20~px to 40~px and then reset, crossed by a white cross 40~px across, with a red center of radius 4~px. The point stayed for 3.1~s. Samples entering the calibration
 fit and its score began 800~ms after the point appeared.

Validation~A and Validation~B used thirteen points. The grid contained the same nine locations, placed at the 10\% insets with no further edge margin, plus four inner points at 30\% and 70\% of screen width and height. Each block drew a new random order of all thirteen points. A point was a static red disc of radius 30~px, a white cross spanning the disc, and a red center of radius 4~px, on white. It stayed for 1.5~s, and the next point followed immediately. The recorded target was the center of the point in pixels of the current viewport.

Between the two validation blocks, participants played a shooting game and were encouraged to move their head.

\subsection{Design}
\label{sec:design}

The study was a single-group, within-participant validation with a fixed order. Every participant received the device check, consent, the device questionnaire, screen-size and camera setup, 9-point calibration with its acceptance check, Validation~A, three rounds of a shooting game, a face-position check, Validation~B, and a debrief. Validation~B repeated Validation~A after the shooting game. The acceptance check used the same 9-point routine as calibration and passed when the mean offset error was at most 3~cm. The setup could be repeated for up to three attempts. A failure to detect the face during the camera check did not use up an attempt. If the face was lost during calibration, the setup returned to the camera check and that restart counted. At the end of the setup the session continued whether or not the criterion had been met, and the outcome was stored. Validation instructions asked the participant to hold the head still and look at the center of each point. Instructions for the shooting game encouraged movement of the head as well as the eyes.

\subsection{Procedure}
\label{sec:procedure}

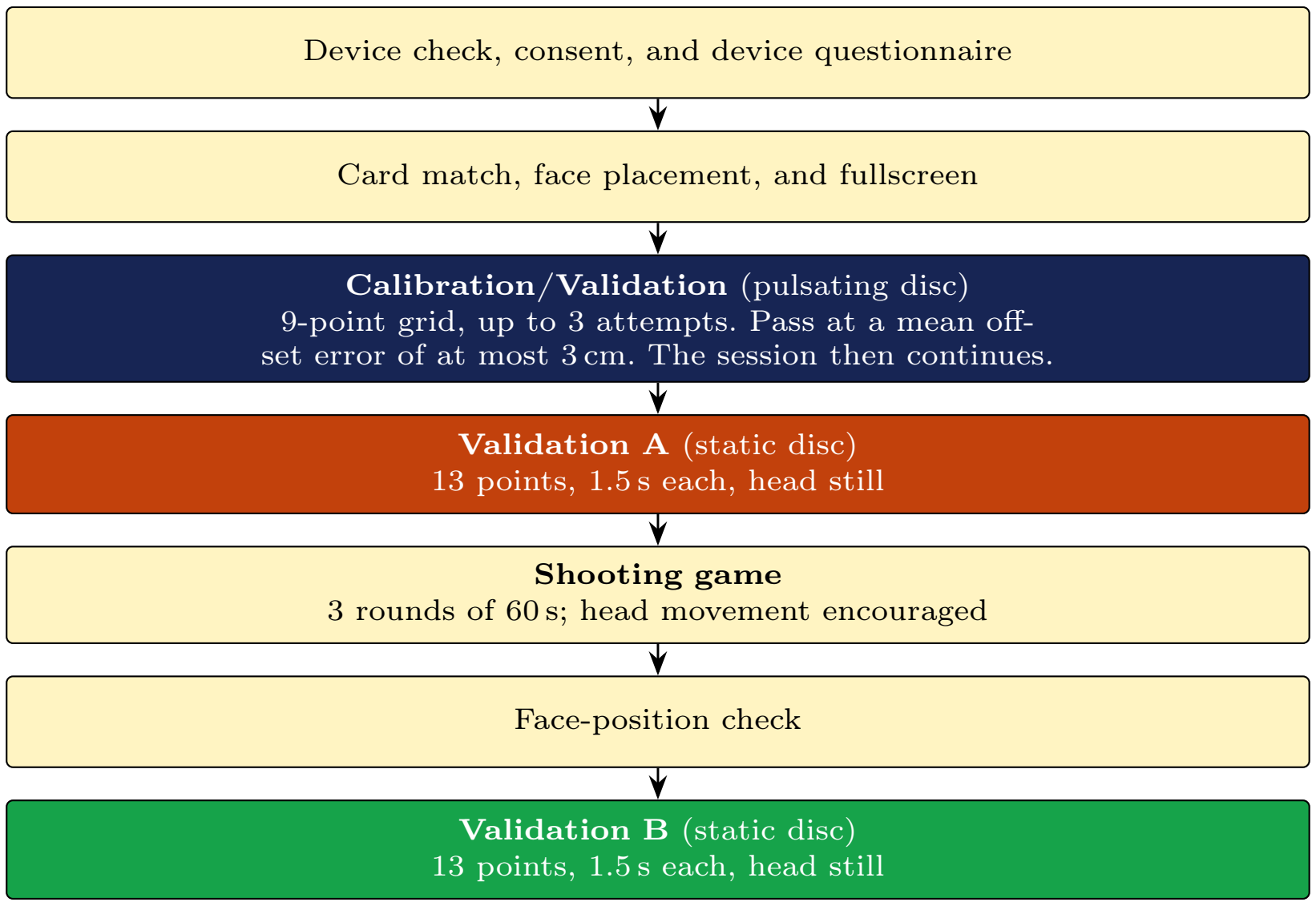
\begin{figure}[H]
  \centering
  \definecolor{procsetup}{RGB}{255,244,194}
  \definecolor{proccal}{HTML}{172554}
  \definecolor{procva}{HTML}{C2410C}
  \definecolor{procvb}{HTML}{16A34A}
  \begin{tikzpicture}[
      font=\scriptsize,
      box/.style={
        draw, rounded corners=1.5pt, align=center,
        inner xsep=5pt, inner ysep=4pt,
        text width=10.4cm, minimum height=7.5mm
      },
      intake/.style={box, fill=procsetup},
      cal/.style={box, fill=proccal, text=white},
      vala/.style={box, fill=procva, text=white},
      valb/.style={box, fill=procvb, text=white},
      arr/.style={-{Stealth[length=2mm]}, line width=0.55pt}
    ]
    \node[intake] (in) {Device check, consent, and device questionnaire};
    \node[intake, below=2.6mm of in] (cam) {Card match, face placement, and fullscreen};
    \node[cal, below=2.6mm of cam] (cal) {%
      \textbf{Calibration/Validation} (pulsating disc)\\[0.5pt]
      9-point grid, up to 3 attempts.
      Pass at a mean offset error of at most 3\,cm.
      The session then continues.};
    \node[vala, below=2.6mm of cal] (va) {%
      \textbf{Validation A} (static disc)\\[0.5pt]
      13 points, 1.5\,s each, head still};
    \node[intake, below=2.6mm of va] (bal) {%
      \textbf{Shooting game}\\[0.5pt]
      3 rounds of 60\,s; head movement encouraged};
    \node[intake, below=2.6mm of bal] (face) {Face-position check};
    \node[valb, below=2.6mm of face] (vb) {%
      \textbf{Validation B} (static disc)\\[0.5pt]
      13 points, 1.5\,s each, head still};
    \draw[arr] (in) -- (cam);
    \draw[arr] (cam) -- (cal);
    \draw[arr] (cal) -- (va);
    \draw[arr] (va) -- (bal);
    \draw[arr] (bal) -- (face);
    \draw[arr] (face) -- (vb);
  \end{tikzpicture}
  \caption{The block order in both the Chiasm and WebGazer experiments.
    Every participant followed this sequence.
    Validation~B repeated Validation~A after the shooting game.}
  \label{fig:procedure}
\end{figure}

After a welcome page, participants gave informed consent and completed the device questionnaire. Setup instructions asked them to sit comfortably and keep the head still. 
They matched the on-screen card to a physical card, made sure their face was centered on the camera image, and entered fullscreen. 
Calibration and the acceptance check followed, each introduced with the instruction to follow each point with the eyes. Validation~A was introduced with the instruction to look at the center of each point and keep the head still. Fullscreen was requested again, and the thirteen points were shown. Participants then read the shooting-game instructions and played three rounds. They were encouraged to move their head during the game. Each round restated the aiming instruction, requested fullscreen, counted down from three, and lasted 60~s, after which a score summary was shown. Before Validation~B the camera image was shown again so the participant could place the face in the oval. A failure of that check was recorded and the session continued into Validation~B, which repeated the thirteen-point procedure. A final page returned the participant to Prolific. Figure~\ref{fig:procedure} shows this order.

\subsection{Data Filtering}
\label{sec:data-filtering}

Gaze is reported in centimetres in the screen plane, using the scale from the card match. Within each validation point, samples from the first 500~ms after onset were discarded. A sample was then discarded when it jumped more than 5~cm from the last kept sample at that point. Exact repeats of a gaze position within the same session and phase were discarded, keeping the first. Samples farther than three robust standard deviations from the median gaze at that point were also discarded. The robust standard deviation is 1.4826 times the median absolute deviation of distances to that median. Points with fewer than five finite samples, and points whose median absolute deviation was zero, were left untrimmed.

\subsection{Evaluation metrics}
\label{sec:metrics}

A person-level value in a phase is the mean of the finite per-point values. It is left undefined when fewer than 9 points on a thirteen-point grid, or fewer than six on the 9-point grid, yield a finite value.

\subsubsection{\textit{Spatial accuracy}}

Spatial accuracy at a point is the median Euclidean distance from gaze to the target. A person's spatial accuracy in a phase is the mean of those point values.
\begin{equation}
\mathrm{accuracy}_{\mathrm{dot}} = \operatorname{median}_{i} \sqrt{(x_{i} - x^{\star})^{2} + (y_{i} - y^{\star})^{2}}
\end{equation}

\subsubsection{\textit{Spatial precision (SD)}}

Spatial precision (SD) at a point is the root-mean-square deviation of the samples from their own centroid. A point with fewer than three finite samples has no spatial precision (SD). A person's spatial precision (SD) in a phase is the mean of the remaining point values.
\begin{equation}
\mathrm{SD}_{\mathrm{dot}} = \sqrt{\mathrm{var}(x) + \mathrm{var}(y)}
\end{equation}

\subsubsection{\textit{Sample-to-sample precision (RMS-S2S)}}

Sample-to-sample precision (RMS-S2S) at a point is the root mean square of the Euclidean distances between successive kept samples, ordered in time. A point with fewer than three finite samples is omitted. A person's sample-to-sample precision (RMS-S2S) in a phase is the mean of the remaining point values.
\begin{equation}
\mathrm{RMS\text{-}S2S}_{\mathrm{dot}} = \sqrt{\frac{1}{T-1}\sum_{t=1}^{T-1}\bigl[(x_{t+1}-x_{t})^{2} + (y_{t+1}-y_{t})^{2}\bigr]}
\end{equation}

\subsubsection{\textit{Drift}}

On the grid, the predicted center at a point is the median, across people, of the median gaze displacement from the target. Drift is the vector from that center on Validation~A to the center on Validation~B. Per person, drift is Validation~B spatial accuracy minus Validation~A spatial accuracy.

\subsubsection{\textit{CDF}}

The cumulative distribution function is the empirical distribution of per-person spatial accuracy, one value per phase. The denominator is the final sample. A person without a finite spatial accuracy stays in that denominator and does not add a step to the curve.

\subsection{Demographics and frame statistics}
\label{sec:chiasm-covariates}

Spatial accuracy on the 9-point validation block was compared across gender, region of residence, and age. Gender was tested with a Welch \(t\)-test of Male versus Female. Region and age were tested with one-way analyses of variance. Age was binned as 18--24, 25--34, 35--44, 45--54, and 55+. A region or age bin entered its analysis only when it contained at least two people. The same accuracy was also compared across the device questionnaire with Welch \(t\)-tests of laptop versus desktop, built-in versus external webcam, and laptop screen versus external monitor. A person with fewer than six usable points was left out of these tests.

The same spatial accuracy was related to the frame statistics on each validation phase. Presence of glasses, eye aspect ratio, face luminance, and contrast were automatically extracted from the frames. Glasses were tested with a \(2\times 3\) analysis of variance, glasses between subjects and validation block within subjects. Eye aspect ratio, face luminance, and Michelson contrast were each tested with a multiple regression of spatial accuracy in centimetres on that measure, validation block, a fixed person intercept, and the measure \(\times\) block interaction.

\subsection{Comparison with WebGazer}
\label{sec:comparison}

We chose the WebGazer implementation in jsPsych \cite{deleeuw2023jspsych}, since it is the most widely used implementation of WebGazer. It uses a 9-point calibration block, during which
participants are asked to click calibration dots, one by one, in a random order, with each location appearing two times. The calibration block is followed 
by a 9-point validation block, during which participants are asked to look at the center of each dot for 2 seconds, with recording starting after 1 second.

The WebGazer dataset was collected using an identical experimental protocol on an independent sample of participants ($N = 101$). The Chiasm dataset was a random 
draw from the included sessions, matched in size to the WebGazer study.
Each dataset used the same procedure for data filtering. Chiasm discarded the first 500~ms after onset on every phase. 
WebGazer kept every sample on the 9-point block and discarded the first 500~ms on Validation~A and Validation~B.

The two datasets were compared on the evaluation metrics. For spatial accuracy, spatial precision (SD), and sample-to-sample precision (RMS-S2S), a planned independent-samples \(t\)-test compared the trackers on each task. A \(2\times 3\) analysis of variance then tested tracker, between subjects, by task, within subjects. A person missing any task was left out of that analysis of variance.

The datasets were compared on demographics and frame statistics. Region of residence, gender, age, and glasses were tested with chi-square tests. 
Face luminance, Michelson contrast, and eye aspect ratio were tested with Welch \(t\)-tests. 

\section{Experimental Results}
\label{sec:results}

We analysed the Chiasm dataset to evaluate the tracker's spatial accuracy, spatial precision (SD), and sample-to-sample precision (RMS-S2S).
Table~\ref{tab:chiasm-metrics} gives the means of each evaluation metric on all three validation blocks. It provides two summaries of the data: 
on with all participants, and the other one with only those who passed the calibration threshold. Further sections describe the data in more detail
and provide statistical tests.

\begin{table}[H]
  \centering
  \caption{Mean evaluation metrics for Chiasm.
    Spatial accuracy, spatial precision (SD), and sample-to-sample precision (RMS-S2S) for all participants and those who passed the calibration 
    threshold.
    $n$ is the number of participants.
    $n$ is smaller in some cells because a participant is omitted when fewer than six points on the validation block, or fewer than nine on Validation~A or Validation~B, yield a finite value for that metric.}
  \label{tab:chiasm-metrics}
  \begin{tabular}{llrrrrrr}
\toprule
 & & \multicolumn{2}{c}{Accuracy} & \multicolumn{2}{c}{Precision (SD)} & \multicolumn{2}{c}{RMS-S2S} \\
\cmidrule(lr){3-4}\cmidrule(lr){5-6}\cmidrule(lr){7-8}
Calibration & Block & $n$ & Mean (cm) & $n$ & Mean (cm) & $n$ & Mean (cm) \\
\midrule
All & validation & 502 & 2.39 & 502 & 1.05 & 502 & 1.29 \\
 & Validation A & 502 & 3.18 & 501 & 1.08 & 501 & 1.34 \\
 & Validation B & 499 & 3.77 & 496 & 1.11 & 496 & 1.36 \\
\midrule
Passed & validation & 460 & 2.15 & 460 & 1.02 & 460 & 1.26 \\
 & Validation A & 457 & 2.96 & 456 & 1.06 & 456 & 1.32 \\
 & Validation B & 455 & 3.53 & 452 & 1.09 & 452 & 1.34 \\
\bottomrule
\end{tabular}

\end{table}

\subsection{Spatial accuracy}
\label{sec:chiasm-spatial}
\begin{figure}[H]
  \centering
  \includegraphics[width=\linewidth,height=0.72\textheight,keepaspectratio]{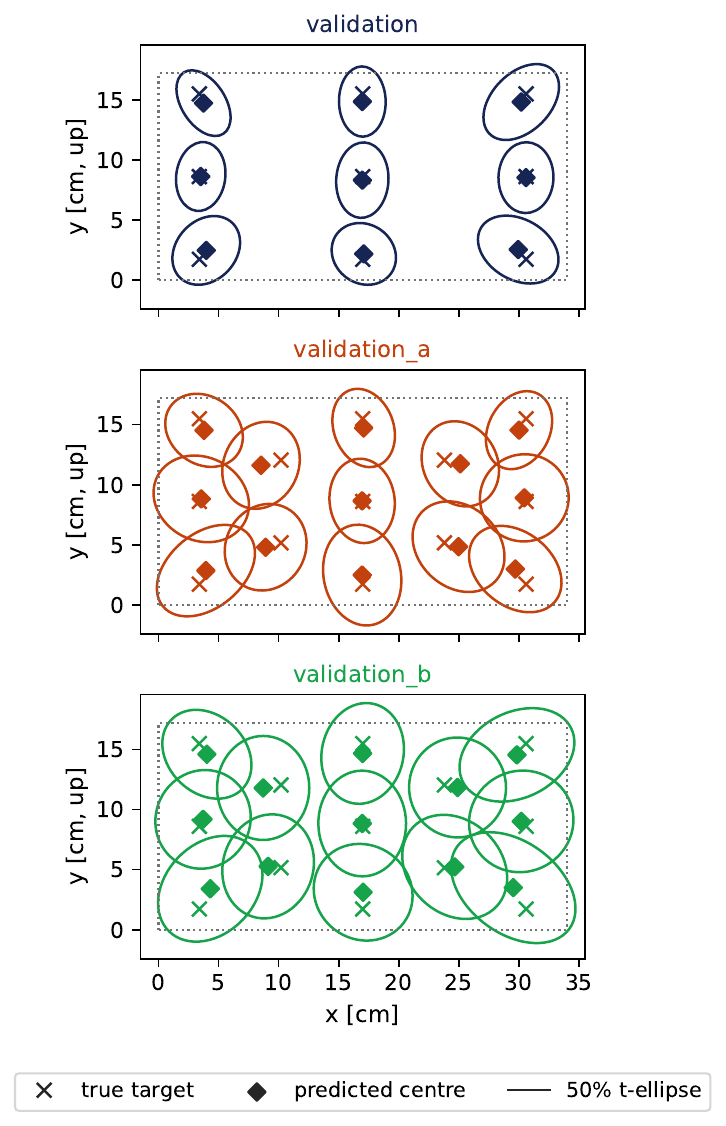}
  \caption{Chiasm spatial accuracy at each true target, with the predicted center marked.}
  \label{fig:02-offset-map}
\end{figure}

\begin{figure}[H]
  \centering
  \includegraphics[width=\linewidth]{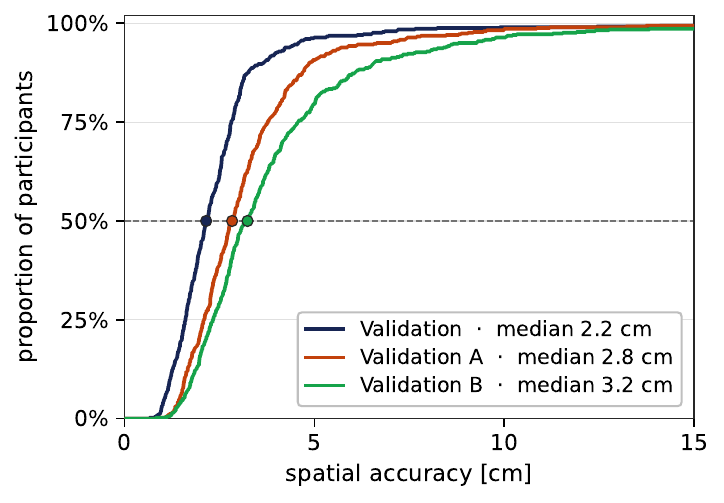}
  \caption{Empirical distribution of Chiasm spatial accuracy.}
  \label{fig:02-ecdf-accuracy}
\end{figure}

In order to visualize the distribution of the typical gaze around the validation points across participants, we computed the median horizontal and
median vertical gaze coordinates for
each point and for each participant (Figure~\ref{fig:02-offset-map}). The predicted center of gaze at each point is the coordinate-wise median of 
these individual offsets (the diamond). The contour is the central 50\% of a bivariate \(t\) distribution with 5 degrees of freedom, located at 
that median and scaled by the covariance of the individual medians. It shows where the typical median gaze landed and how that landing place varied 
across participants.
The median of the predicted-to-target distances was 0.64~cm on the validation block (9-point), 1.12~cm on Validation~A (13 points), and 1.09~cm on
Validation~B (13 points). Both later blocks were farther from the target than the validation block, and Validation~A and Validation~B were similar.
The contours, the spread of the predicted center across participants, were larger on Validation~A than on the validation block and larger again on
Validation~B. At the points at the edges of the screen, the predicted centers lay inward of the targets, toward the screen center. The contour tilt
means participants differ most along the center-to-edge direction, which may be due to regression to the mean for extreme positions.
The empirical distribution of spatial accuracy is shown in Figure~\ref{fig:02-ecdf-accuracy} and give a clear picture of the distribution of 
spatial accuracy across participants. Each person contributes one value per phase: the mean
of that person's per-dot median distances to the target. The curve is the share of those participants whose value is at or below each distance. 
There must be at least six usable points on the validation block and at least nine on Validation~A and Validation~B in order to be included.
The mean was 2.39~cm on the validation block, 3.18~cm on Validation~A, and 3.77~cm on Validation~B. 
A repeated-measures analysis of variance of validation block, using the 494 people with a value on every block, was reliable, \(F(2, 986) = 162.00\),
\(p < 0.001\). Planned paired \(t\)-tests distinguished the validation block from Validation~A, \(t = -12.28\), \(df = 498\), \(p < 0.001\), and 
Validation~A from Validation~B, \(t = -7.52\), \(df = 496\), \(p < 0.001\). This shows a systematic decrease in spatial accuracy on the later blocks.

\subsection{Spatial precision (SD)}
\label{sec:chiasm-precision}
\begin{figure}[H]
  \centering
  \includegraphics[width=\linewidth,height=0.72\textheight,keepaspectratio]{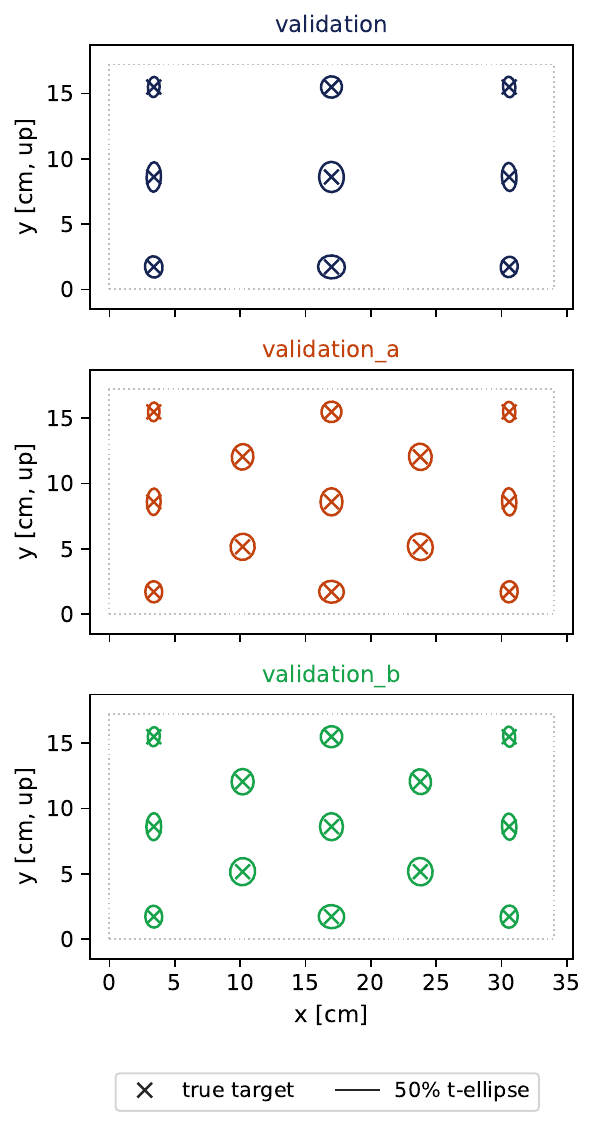}
  \caption{Chiasm within-person spatial precision (SD) drawn at each true target.}
  \label{fig:02-precision-map}
\end{figure}

In order to visualize the typical within-person spread of gaze at each validation point, we computed the covariance of the samples around that point's own centroid for each point and for each participant (Figure~\ref{fig:02-precision-map}). Spatial precision (SD) is the root mean square of that spread. The contour is the central 50\% of a bivariate \(t\) distribution with 5 degrees of freedom, centered on the true target and scaled by the median of these covariances across participants. It shows the typical size and tilt of the within-person scatter, separately from where the gaze landed.
The median spatial precision (SD), across participants and points, was 0.95~cm on the validation block (9-point), 0.97~cm on Validation~A (13 points), and 0.99~cm on Validation~B (13 points). The later blocks were only slightly less precise than the validation block. The contours extend farther vertically than horizontally, so spatial precision seems better in the horizontal dimension than in the vertical.
Each person then contributes one value per phase: the mean of that person's per-dot precision values. There must be at least six usable points on the validation block and at least nine on Validation~A and Validation~B in order to be included.
The mean was 1.05~cm on the validation block, 1.08~cm on Validation~A, and 1.11~cm on Validation~B. A repeated-measures analysis of variance of validation block, using the 490 people with a value on every block, was reliable, \(F(2, 978) = 19.86\), \(p < 0.001\). Planned paired \(t\)-tests distinguished the validation block from Validation~A, \(t = -4.15\), \(df = 497\), \(p < 0.001\), and Validation~A from Validation~B, \(t = -2.81\), \(df = 492\), \(p = 0.005\). This shows a small systematic decrease in spatial precision on the later blocks.

\subsection{Sample-to-sample precision (RMS-S2S)}
\label{sec:chiasm-s2s}
In order to measure how far gaze moved from one sample to the next, we computed the root mean square of the Euclidean distances between successive kept samples at each point and for each participant. Each person then contributes one value per phase: the mean of that person's per-dot values. There must be at least six usable points on the validation block and at least nine on Validation~A and Validation~B in order to be included.
The mean was 1.29~cm on the validation block, 1.34~cm on Validation~A, and 1.36~cm on Validation~B. A repeated-measures analysis of variance of validation block, using the 490 people with a value on every block, was reliable, \(F(2, 978) = 23.17\), \(p < 0.001\). Planned paired \(t\)-tests distinguished the validation block from Validation~A, \(t = -6.01\), \(df = 497\), \(p < 0.001\). Validation~A and Validation~B did not differ, \(t = -1.75\), \(df = 492\), \(p = 0.080\). This shows a small decrease in sample-to-sample precision from the validation block to Validation~A, and no further change on Validation~B.

\subsection{Drift}
\label{sec:chiasm-drift}
\begin{figure}[H]
  \centering
  \includegraphics[width=\linewidth]{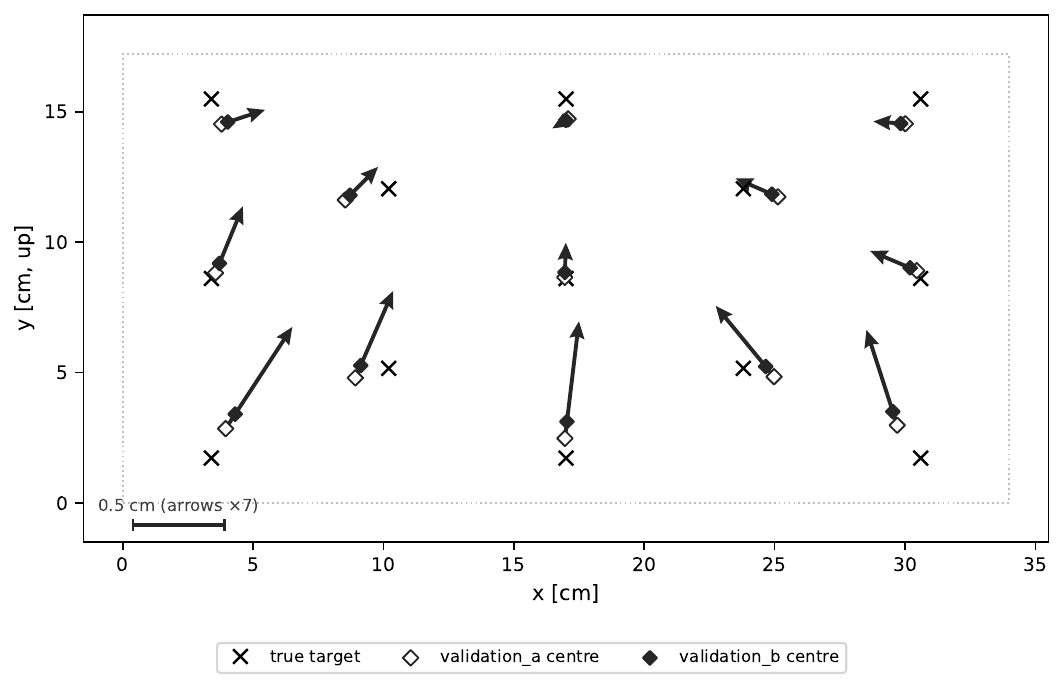}
  \caption{Chiasm drift on the dot grid.
    Each arrow runs from the validation~A predicted center to the validation~B center.
    The arrows are drawn at seven times their length for better readability.}
  \label{fig:02-drift-dot}
\end{figure}

In order to visualize how the predicted center moved from Validation~A to Validation~B, we drew one arrow at each of the 13 validation points (Figure~\ref{fig:02-drift-dot}). The tail is the predicted center on Validation~A and the head is the predicted center on Validation~B, the same center as the diamond in Figure~\ref{fig:02-offset-map}. The arrows are drawn at seven times their length for better readability. All 13 points had at least nine participants on both blocks (504 on Validation~A and 501 on Validation~B).
The mean shift across these points was \(-0.00\)~cm horizontally and 0.27~cm upward. The 0.37~cm in the abstract is the mean length of these shifts: at each of the 13 points, the distance from the Validation~A predicted center to the Validation~B predicted center, then the average of those 13 distances. A one-sample \(t\)-test found no reliable horizontal shift, \(t = -0.02\), \(df = 12\), \(p = 0.99\). The upward shift was reliable, \(t = 4.34\), \(df = 12\), \(p < 0.001\). This shows a small upward shift of the predicted center, and no horizontal shift.

\subsection{Devices}
\label{sec:chiasm-devices}

The participants who took part in the study used primarily laptops (439) and desktops (66), with a built-in webcam (429) or an external one (75). 
They typically used laptop display (429), and some used an external monitor (64), or an all-in-one setup (12).
Figure~\ref{fig:01-device-questionnaire} shows the device questionnaire for the 505 included participants. 

\subsection{Demographics}
\label{sec:chiasm-demographics}

The data from the Chiasm study comes from a diverse population. Of the 504 participants with a country of residence, 199 lived in Europe, 107 in Western, Central and 
Southern Asia, 67 in Africa, 56 in Latin America and the Caribbean, 28 in North America, 22 in Eastern and 
South-Eastern Asia, 20 in the Middle East, and 5 in Oceania. Figure~\ref{fig:02-demographics} in Appendix
shows region of residence, gender, and age.
Furthermore, we found that demographic information had a limited effect on on spatial accuracy in the 
validation block (see Figures~\ref{fig:02-demographics-accuracy-sex}, 
\ref{fig:02-demographics-accuracy-region}, and~\ref{fig:02-demographics-accuracy-age} 
in Appendix~\ref{app:chiasm-figures}.
). Men (\(n = 310\)) had a mean of 2.50~cm and women 
(\(n = 191\)) a mean of 2.20~cm. A Welch \(t\)-test distinguished them, \(t = 2.49\), \(p = 0.013\). 
Region of residence and age showed no reliable difference, \(F(7, 493) = 1.26\), \(p = 0.27\), and 
\(F(4, 496) = 1.09\), \(p = 0.36\).
Laptop users (\(n = 436\)) had a mean of 2.31~cm and desktop users (\(n = 66\)) a mean of 2.88~cm. A Welch \(t\)-test distinguished them, \(t = -2.32\), \(p = 0.023\).
Built-in webcams (\(n = 427\)) had a mean of 2.32~cm and external webcams (\(n = 74\)) a mean of 2.75~cm. That difference was not reliable, \(t = -1.95\), \(p = 0.054\).
Laptop screens (\(n = 426\)) had a mean of 2.29~cm and external monitors (\(n = 64\)) a mean of 2.87~cm. A Welch \(t\)-test distinguished them, \(t = -3.16\), \(p = 0.002\).

\subsection{Frame statistics}
\label{sec:chiasm-frames}

In order to relate the camera image to spatial accuracy, we tested glasses, eye aspect ratio, face luminance, and Michelson contrast
(see Figures~\ref{fig:02-frame-glasses}--\ref{fig:02-frame-contrast} in Appendix~\ref{app:chiasm-figures}).
Forty-two participants wore glasses and 463 did not. Glasses decreased spatial accuracy, \(F(1, 503) = 13.50\), \(p < 0.001\). The offset was 2.90 versus 2.36~cm on the validation block, 4.06 versus 3.11~cm on Validation~A, and 4.95 versus 3.73~cm on Validation~B. The glasses \(\times\) block interaction was not reliable, \(F(2, 1006) = 2.84\), \(p = 0.059\) (Figure~\ref{fig:02-frame-glasses}).
A higher eye aspect ratio, a more open eye, was associated with a smaller offset, slope \(-7.25\)~cm per unit, \(F(1, 1005) = 4.84\), \(p = 0.028\). Observed eye aspect ratio ranged from 0.18 to 0.40, which is about 1.6~cm on this slope.

\subsection{Comparison with WebGazer}
\label{sec:comparison-results}

Demographics and frame statistics did not differ between the Chiasm subsample and the WebGazer study, except for Michelson contrast, which was higher for WebGazer.
Region of residence, gender, and age showed no reliable difference, \(\chi^{2}(7) = 6.68\), \(p = 0.46\), \(\chi^{2}(1) = 0.08\), \(p = 0.77\), and \(\chi^{2}(4) = 2.67\), \(p = 0.61\).
Face luminance, eye aspect ratio, and glasses did not differ reliably either, \(t = -1.62\), \(p = 0.11\), \(t = 0.82\), \(p = 0.41\), and \(\chi^{2}(1) = 0.08\), \(p = 0.78\).
Michelson contrast was \(-0.05\) for WebGazer and \(-0.14\) for Chiasm, \(t = -2.92\), \(p = 0.004\).

The comparison maps for spatial accuracy, spatial precision (SD), and drift are shown in Figures~\ref{fig:06-offset-map}, 
\ref{fig:06-precision-map}, and~\ref{fig:06-drift-dot} in Appendix~\ref{app:comparison-figures}. It can clearly be seen that the spatial accuracy and spatial precision (SD)
of Chiasm are better than that of WebGazer across all validation blocks. For WebGazer, the oblong shape of the spatial accuracy contours, tilted towards the center of 
the screen, suggests that the participants tend to regress to the mean for extreme positions. The shape of the spatial precision (SD) contours was stretched in the
horizontal dimension for WebGazer, but in vertical dimension for Chiasm. The mean drift across the 13 points was 0.47~cm for Chiasm and 0.68~cm for WebGazer. A paired \(t\)-test did not distinguish them, \(t = -1.64\), \(df = 12\), \(p = 0.13\). The Chiasm arrows pointed upward and toward the center of the screen. For WebGazer, the drift appeared to be random. 

\begingroup
\makeatletter
\renewcommand\subsubsection{\@startsection{subsubsection}{3}{\z@}%
  {-12\p@ \@plus -4\p@ \@minus -4\p@}%
  {6\p@ \@plus 2\p@ \@minus 2\p@}%
  {\normalfont\normalsize\bfseries\boldmath
   \rightskip=\z@ \@plus 8em\pretolerance=10000 }}
\makeatother

\subsubsection{Evaluation metrics}
\label{sec:comparison-ecdf}

The empirical distribution of spatial accuracy for Chiasm and WebGazer is shown in Figure~\ref{fig:06-ecdf-accuracy} and give a clear picture of the distribution of 
spatial accuracy across participants. It is clear that Chiasm outperforms WebGazer across all validation blocks.
\begin{figure}[H]
  \centering
  \includegraphics[width=\linewidth]{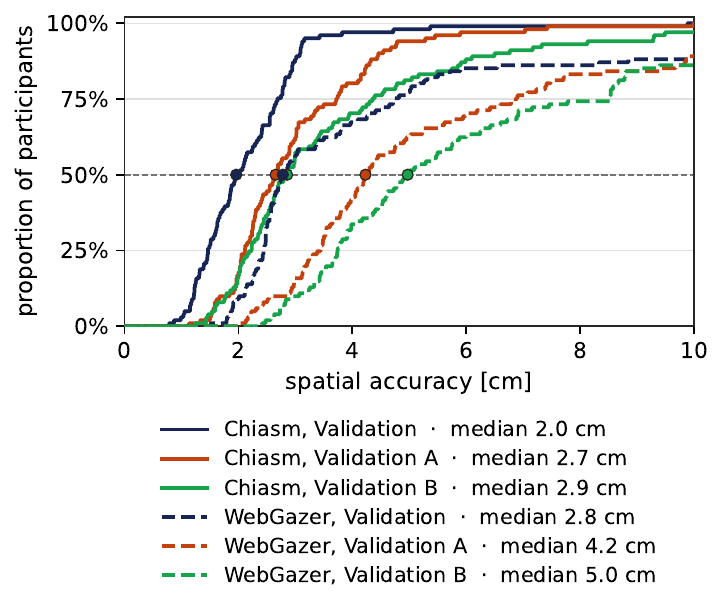}
  \caption{Spatial accuracy distributions.
    Solid lines are Chiasm and dashed lines are WebGazer.
    Colour is the validation phase.}
  \label{fig:06-ecdf-accuracy}
\end{figure}

Table~\ref{tab:comparison-metrics} reports the means of each evaluation metric for the Chiasm subsample and the WebGazer study. 

\begin{table}[H]
  \centering
  \caption{Mean evaluation metrics for the Chiasm subsample and the WebGazer study.
    Spatial accuracy, spatial precision (SD), and sample-to-sample precision (RMS-S2S).
    The Chiasm rows are the random subsample matched in size to the WebGazer study.
    $n$ is the number of participants.
    $n$ is smaller in some cells because a participant is omitted when fewer than six points on the validation block, or fewer than nine on Validation~A or Validation~B, yield a finite value for that metric.}
  \label{tab:comparison-metrics}
  \begin{tabular}{llrrrrrr}
\toprule
 & & \multicolumn{2}{c}{Accuracy} & \multicolumn{2}{c}{Precision (SD)} & \multicolumn{2}{c}{RMS-S2S} \\
\cmidrule(lr){3-4}\cmidrule(lr){5-6}\cmidrule(lr){7-8}
Study & Block & $n$ & Mean (cm) & $n$ & Mean (cm) & $n$ & Mean (cm) \\
\midrule
Chiasm & validation & 101 & 2.18 & 101 & 1.01 & 101 & 1.26 \\
 & Validation A & 101 & 3.01 & 101 & 1.05 & 101 & 1.33 \\
 & Validation B & 100 & 3.62 & 100 & 1.09 & 100 & 1.36 \\
\midrule
WebGazer & validation & 101 & 4.51 & 96 & 1.55 & 96 & 0.79 \\
 & Validation A & 98 & 5.62 & 94 & 1.61 & 94 & 0.91 \\
 & Validation B & 97 & 6.18 & 91 & 1.64 & 91 & 0.95 \\
\bottomrule
\end{tabular}

\end{table}

\makeatletter
\renewcommand\subsubsection{\@startsection{subsubsection}{3}{\z@}%
  {-18\p@ \@plus -4\p@ \@minus -4\p@}%
  {-0.5em \@plus -0.22em \@minus -0.1em}%
  {\normalfont\normalsize\itshape}}
\makeatother

\subsubsection{Mean spatial accuracy}%
\label{sec:comparison-mean-accuracy}%
Figure~\ref{fig:06-mean-metrics}(a) shows one spatial-accuracy value per person. A \(2 \times 3\) analysis of variance, with tracker between
subjects and validation block within subjects, used the 100 Chiasm and 97 WebGazer participants with a value on every block. 
The tracker effect was reliable, \(F(1, 195) = 35.03\), \(p < 0.001\), as was the block effect, \(F(2, 390) = 66.14\), \(p < 0.001\). 
The tracker \(\times\) block interaction was not reliable, \(F(2, 390) = 1.70\), \(p = 0.18\). 
This shows better spatial accuracy for Chiasm on every block.

\subsubsection{Mean spatial precision (SD)}%
\label{sec:comparison-mean-precision}%
Figure~\ref{fig:06-mean-metrics}(b) shows one spatial-precision value per person. 
The tracker effect was reliable, \(F(1, 189) = 71.52\), \(p < 0.001\), as was the block effect, \(F(2, 378) = 12.27\), \(p < 0.001\). 
The tracker \(\times\) block interaction was not reliable, \(F(2, 378) = 0.64\), \(p = 0.53\). 
This shows better spatial precision for Chiasm than WebGazer on every block.

\subsubsection{Mean sample-to-sample precision (RMS-S2S)}%
\label{sec:comparison-mean-s2s}%
Figure~\ref{fig:06-mean-metrics}(c) shows one sample-to-sample value per person. The tracker effect was reliable, \(F(1, 189) = 58.29\), \(p < 0.001\), 
as was the block effect, \(F(2, 378) = 65.96\), \(p < 0.001\). The tracker \(\times\) block interaction was also reliable,
 \(F(2, 378) = 9.38\), \(p < 0.001\). Chiasm had the larger RMS on every block, so its sample-to-sample precision was lower, and that difference decreased slightly 
over time.

\begin{figure}[H]
  \centering
  \includegraphics[width=\linewidth]{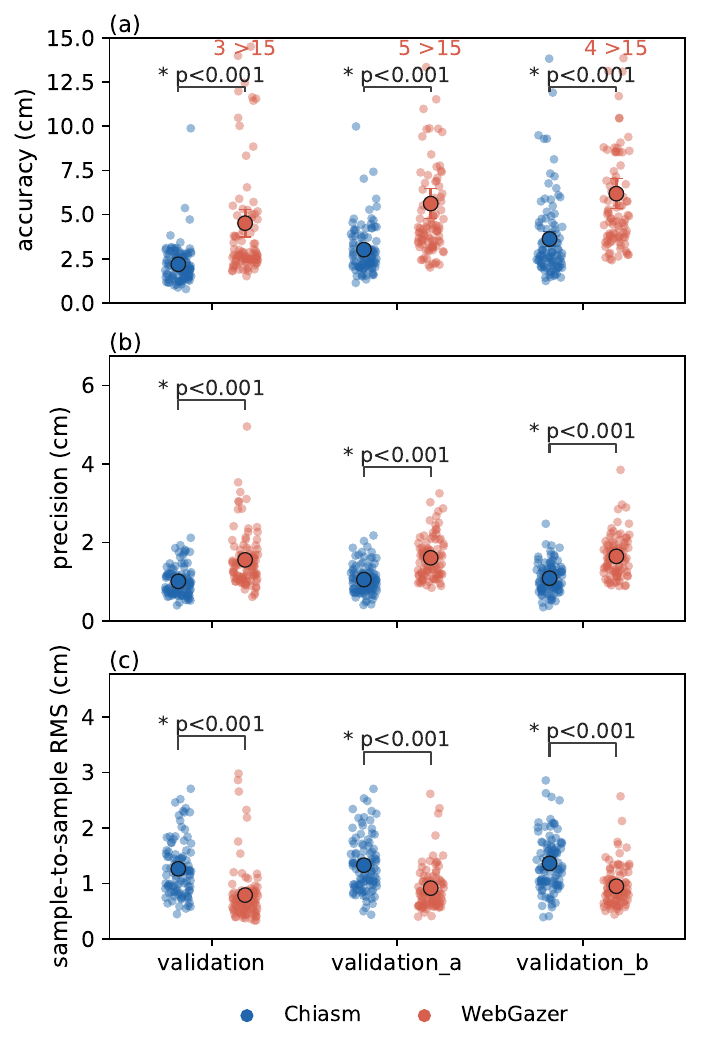}
  \caption{Person-level means for the Chiasm subsample and the WebGazer study.
    (a)~Spatial accuracy.
    (b)~Spatial precision (SD).
    (c)~Sample-to-sample precision (RMS-S2S).
    Each person is a point.
    The marker is the mean with its 95\% confidence interval, and the bracket is an independent-samples \(t\)-test.
    Counts above 15~cm stay in the mean and the test.}
  \label{fig:06-mean-metrics}
\end{figure}

\endgroup

\section{Discussion}
\label{sec:discussion}

The results from our large-scale validation study ($N = 505$) show that Chiasm performs better than existing solutions ``in the wild'', where the experimenter has no control over the devices, environment or 
the participant's behavior. The participants came from a diverse population in terms of region of residence, gender, age, and used a variety of devices. Our comprehensive
analysis showed that Chiasm has a spatial accuracy (2.39~cm) and precision (SD 1.05~cm), and the sample-to-sample 
precision (1.29~cm) is also reasonable given the 30Hz sampling rate. These evaluation metrics improved slightly when only the participants who passed the calibration 
were included in the analysis (see Table~\ref{tab:chiasm-metrics}). The spatial accuracy and precision (SD) decreased over time, but the sample-to-sample precision
was more stable. The drift over time was small and directed upward and toward the center of the screen, which is consistent with regression to the center of the screen
\cite{saxena2024}.
Interestingly, the spatial accuracy decreased sharply from the validation block to Validation~A, but only slightly to Validation~B. This is likely due to the switch from
using dynamic to static targets. In validation block, the targets were pulsating discs, which captured participants' attention in automatic fashion. In Validation~A 
and Validation~B, the same discs, but static, which required more effort from the participants to track. While such static discs are the most common type of calibration
targets, they are probably not the most effective way to enforce fixation. Given that participants were not trained in eye-tracking tasks,
and did not receive feedback from the experimenter, this is not surprising that the spatial accuracy decreased.

In terms of external variables affecting spatial accuracy, the presence of glasses and the lower eye aspect ratio had substantial detrimental effects. The face luminance
or contrast had no effect on spatial accuracy. The demographic variables also had no effect on spatial accuracy, except for gender. 
Chiasm's spatial accuracy is comparable to the iMotions white paper report (median of $2.1^{\circ}$), which is the only other large-scale validation study conducted
``in the wild'' ($N = 255$). They also found the effect of glasses to be detrimental. Unfortunately, the iMotions study did not report other metrics and were not clear about
the exact experimental setup. 

Chiasm also outperformed the FAZE model, which included user-specific calibration \cite{saxena2024}, but was run fully offline, on both spatial
accuracy of $2.28^{\circ}$ vs $2.44^{\circ}$ and a spatial precision (SD) of $1.00^{\circ}$ vs $1.63^{\circ}$. Since the dataset on which FAZE was evaluated
excluded participants with glasses, Chiasm's advantage over FAZE may be underestimated. FAZE yielded a lower RMS-S2S than our system 
($0.47^{\circ}$ vs.\ $1.23^{\circ}$). This difference likely reflects offline temporal smoothing of the gaze estimates by the Kalman filter in the 
FAZE pipeline. In addition, Chiasm runs online with a near real-time inference speed, while FAZE requires fully offline processing. Interestingly, 
in that study \cite{saxena2024}, running the models fully \textit{offline} on webcam frames collected \textit{online}, resulted in high attrition, 
since only 65 out of 118 participants were included in the analysis. This illustrates a limitation of fully offline processing, and emphasizes the importance
 of user interface design for online gaze tracking. 

The results from the comparison with WebGazer ($N = 101$) show that Chiasm performs substantially better than WebGazer across all validation blocks 
for spatial accuracy, spatial precision (SD), but not for sample-to-sample precision (RMS-S2S) (see Table~\ref{tab:comparison-metrics}). In combination with 
poor spatial accuracy than Chiasm, better sample-to-sample precision is likely to reflect the consequences of the ridge regression algorithm used by WebGazer. It
suggests that the algorithm is taking smaller steps than Chiasm, but these steps are less clustered around the target.

Overall, we present a rigorous, comprehensive and extensive evaluation of Chiasm's performance ``in the wild'', as well as its comparison with WebGazer. Besides evaluating
it on all standard metrics such as spatial accuracy, spatial precision (SD), and sample-to-sample precision (RMS-S2S), we also evaluated it on drift of these metrics over
time, which gives a realistic picture of the tracker's performance in a real-world scenario.

\section{Conclusion}
\label{sec:conclusion}


In this study, which to our knowledge is the largest validation of a webcam-based
gaze tracker under unconstrained ``in the wild'' conditions to date
($N = 505$), Chiasm achieved a mean spatial accuracy of approximately
$2.3^{\circ}$ of visual angle. This outperforms WebGazer, which we evaluated
directly, and compared favourably with previously reported webcam-based accuracies. 
Chiasm thus narrows the gap to laboratory-based eye trackers: remote video-based trackers achieve about $0.5^{\circ}$ under optimal
conditions, and wearable eye trackers typically around $1.5^{\circ}$
\cite{ehinger2019}. A direct comparison with a laboratory
eye tracker under controlled conditions, ideally with simultaneous recording,
will be an important next step. We are confident that Chiasm is 
ready to be used in a variety of real-world tasks, relying on multiple areas of interest (AOIs),
that are currently mainly performed in the lab. Chiasm is now available as an extension to jsPsych \cite{deleeuw2023jspsych}, making it easy to use in a wide 
range of research and applications.

\section*{Funding}

This study was funded by Chiasm B.V.

\section*{Competing Interests}

A.B. and F.P\ are co-founders and shareholders of Chiasm B.V., which develops the Chiasm software evaluated here.

\section*{Data Availability}

Data is made available at [Zenodo link].

\printbibliography

\clearpage
\appendix

\section{Attrition in the Chiasm study}
\label{app:chiasm-attrition}

\begin{figure}[H]
  \centering
  \includegraphics[width=0.92\linewidth]{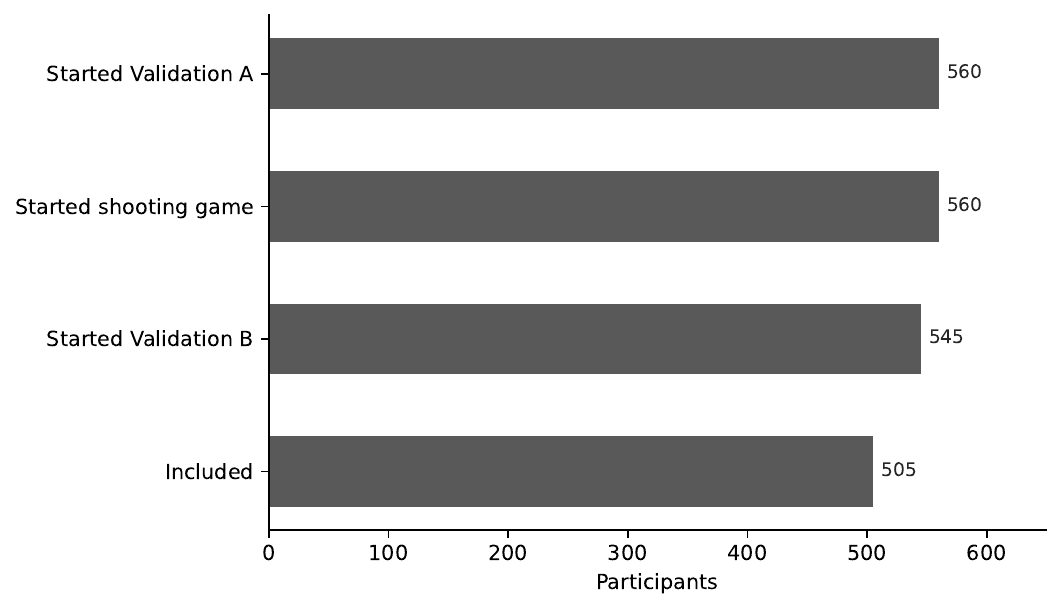}
  \caption{Attrition after the start of Validation~A in the Chiasm study.
    Each bar counts participants who started that stage.    
    Included are the participants who contributed to all validation blocks.    
    }
  \label{fig:chiasm-attrition}
\end{figure}

\section{Attrition in the WebGazer study}
\label{app:attrition}

\begin{figure}[H]
  \centering
  \includegraphics[width=0.92\linewidth]{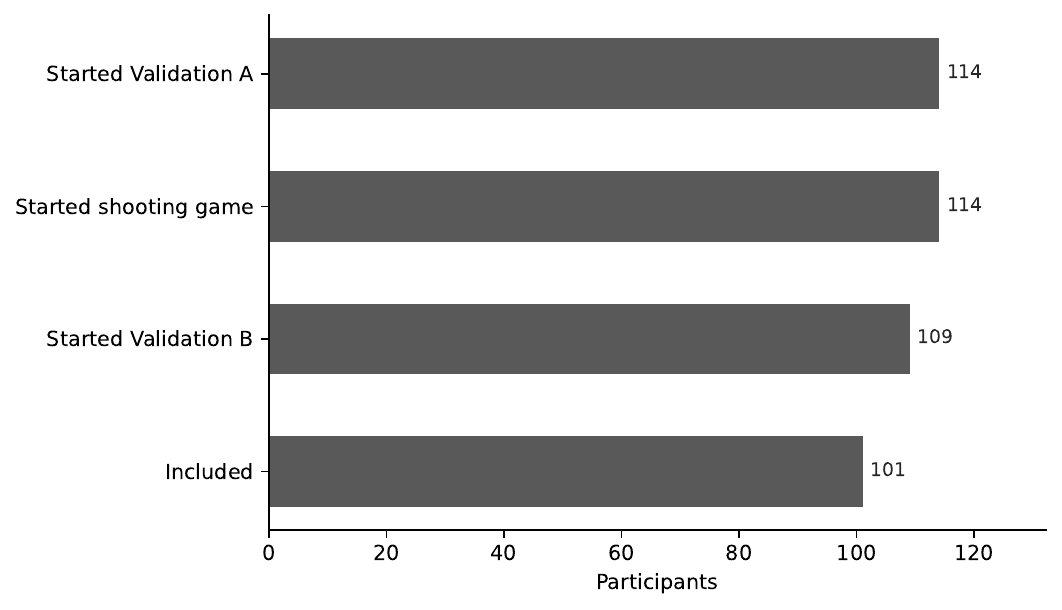}
  \caption{Attrition after the start of Validation~A in the WebGazer study.
    Each bar counts participants who started that stage.    
    Included are the participants who contributed to all validation blocks.}
  \label{fig:webgazer-attrition}
\end{figure}

\section{Chiasm benchmark figures}
\label{app:chiasm-figures}

\begin{figure}[H]
  \centering
  \includegraphics[width=\linewidth]{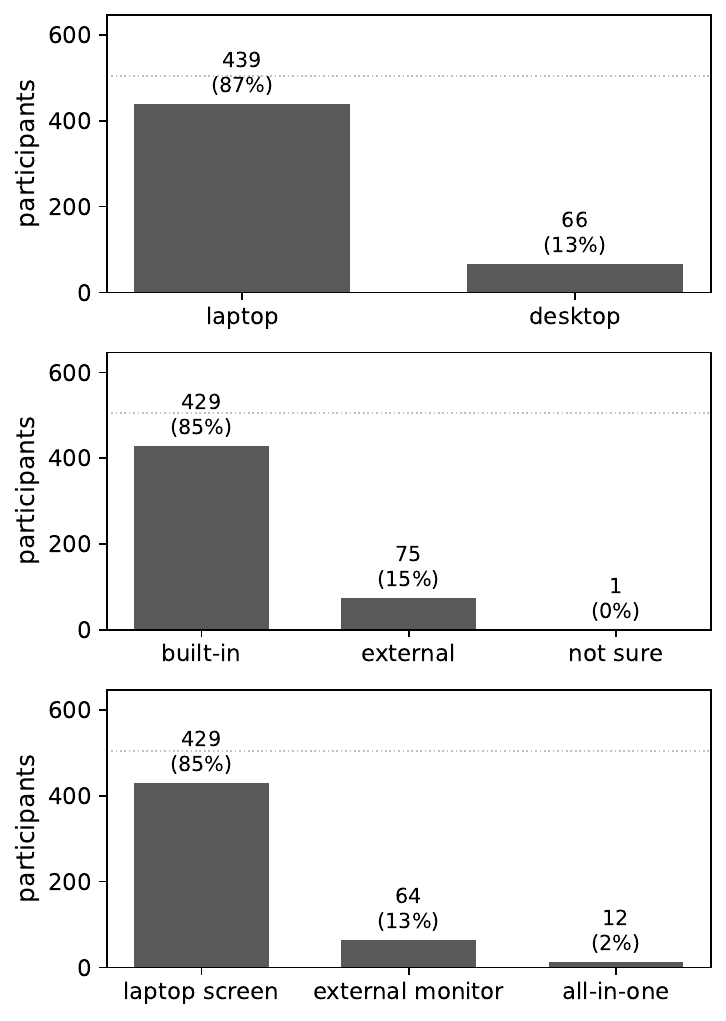}
  \caption{Computer, camera, and display reported by the 505 included participants.}
  \label{fig:01-device-questionnaire}
\end{figure}

\begin{figure}[H]
  \centering
  \includegraphics[width=\linewidth]{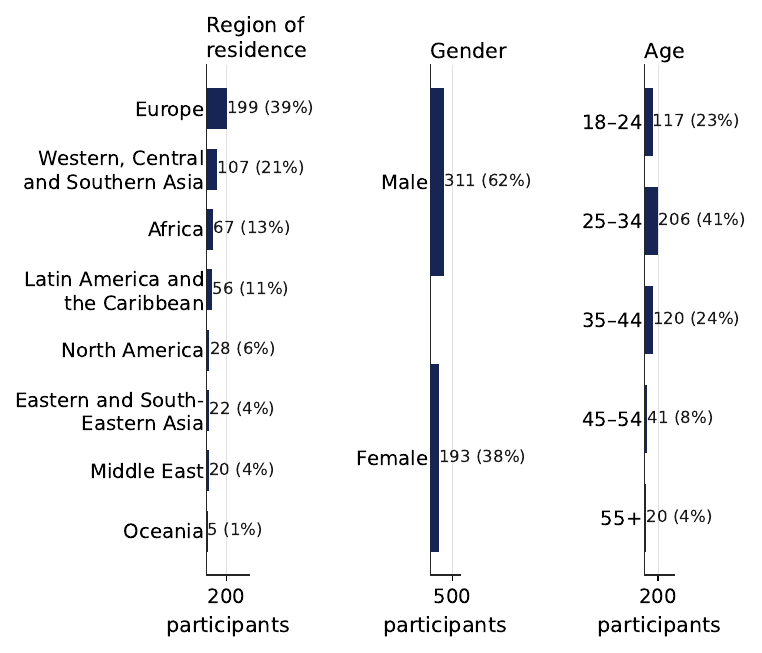}
  \caption{Region of residence, gender, and age in the Chiasm included sample.
    A missing Prolific field is left off that panel.}
  \label{fig:02-demographics}
\end{figure}

\begin{figure}[H]
  \centering
  \includegraphics[width=\linewidth]{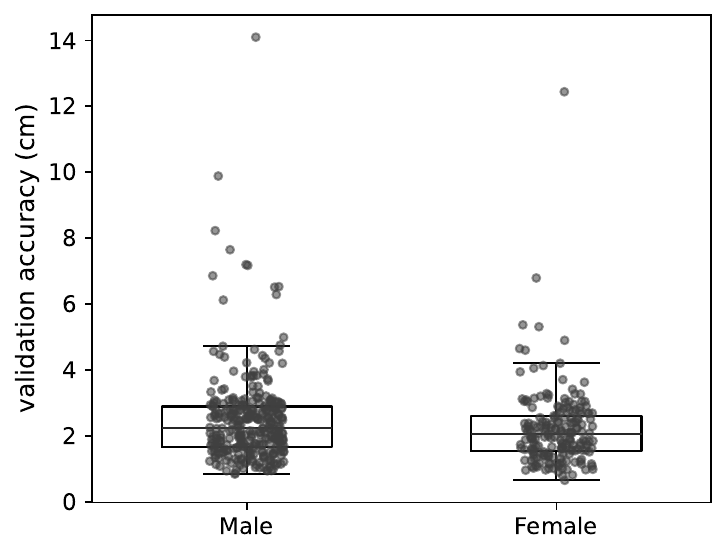}
  \caption{Chiasm validation spatial accuracy by gender.
    A person with fewer than six usable dots is omitted.}
  \label{fig:02-demographics-accuracy-sex}
\end{figure}

\begin{figure}[H]
  \centering
  \includegraphics[width=\linewidth]{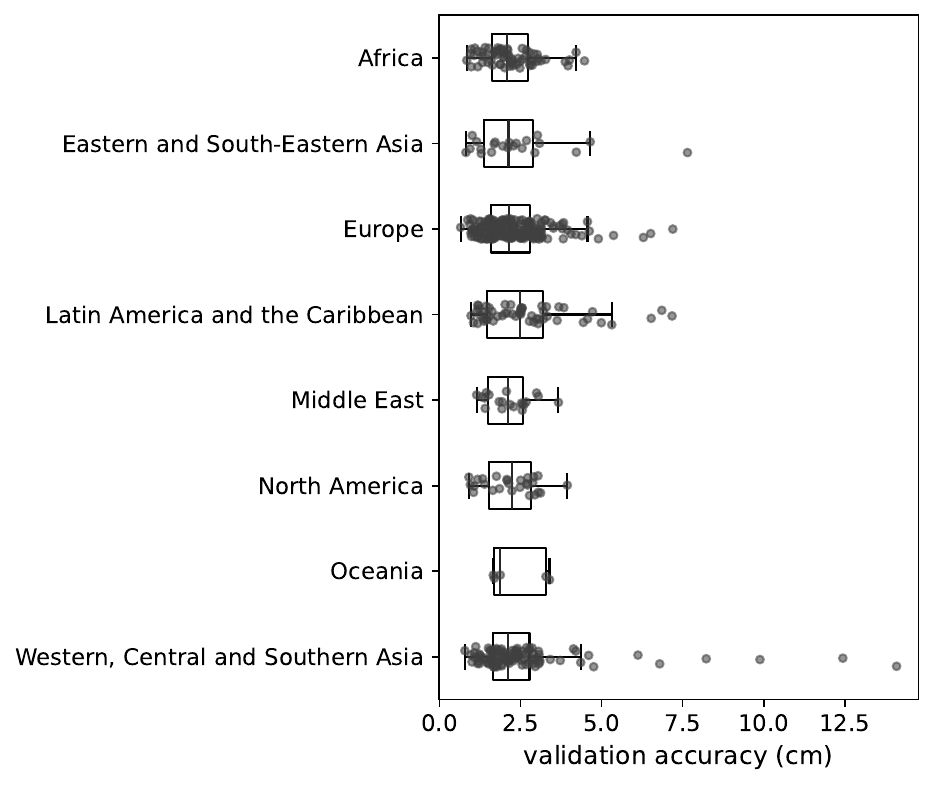}
  \caption{Chiasm validation spatial accuracy by region of residence.
    A person with fewer than six usable dots is omitted.}
  \label{fig:02-demographics-accuracy-region}
\end{figure}

\begin{figure}[H]
  \centering
  \includegraphics[width=\linewidth]{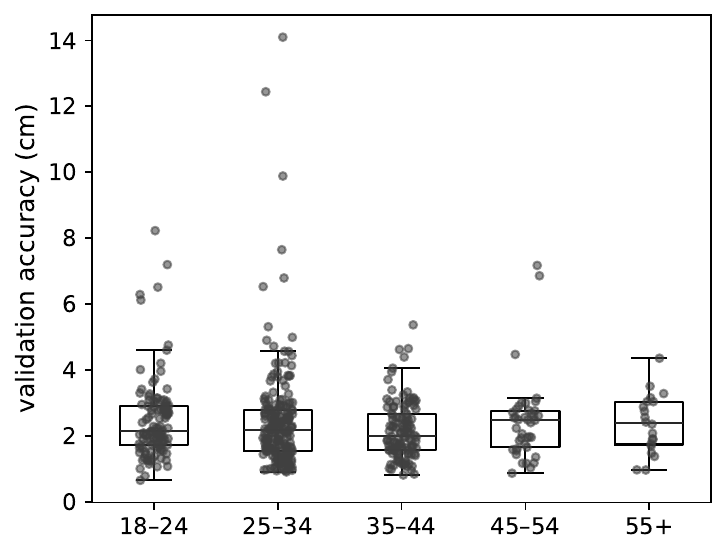}
  \caption{Chiasm validation spatial accuracy by age.
    A person with fewer than six usable dots is omitted.}
  \label{fig:02-demographics-accuracy-age}
\end{figure}

\begin{figure}[H]
  \centering
  \includegraphics[width=\linewidth]{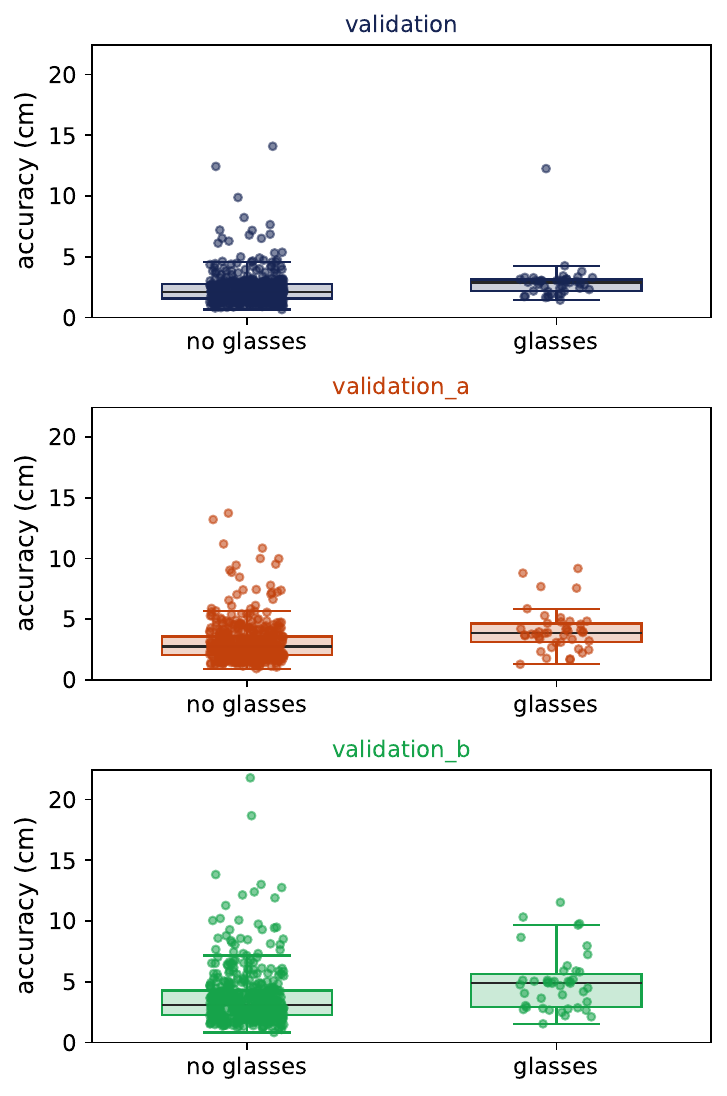}
  \caption{Chiasm mean spatial accuracy by glasses.
    Glasses is the session median of the face-frame glasses score, at least one half.}
  \label{fig:02-frame-glasses}
\end{figure}

\begin{figure}[H]
  \centering
  \includegraphics[width=\linewidth]{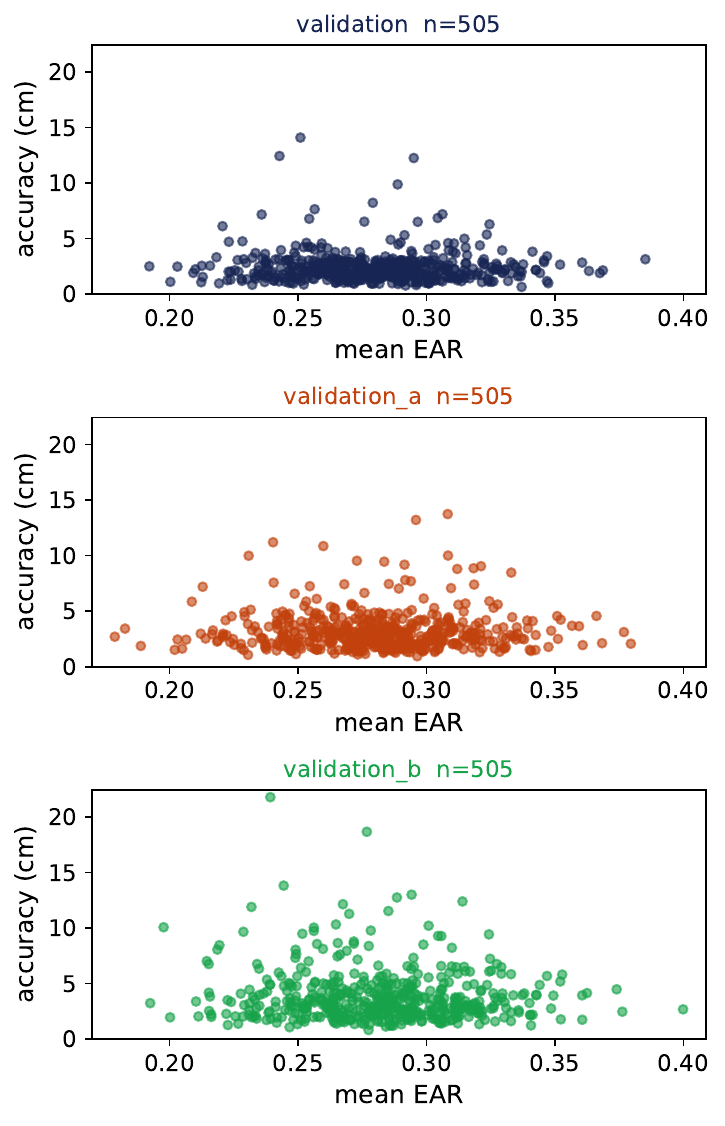}
  \caption{Chiasm mean spatial accuracy against eye aspect ratio on the same validation phase.}
  \label{fig:02-frame-ear}
\end{figure}

\begin{figure}[H]
  \centering
  \includegraphics[width=\linewidth]{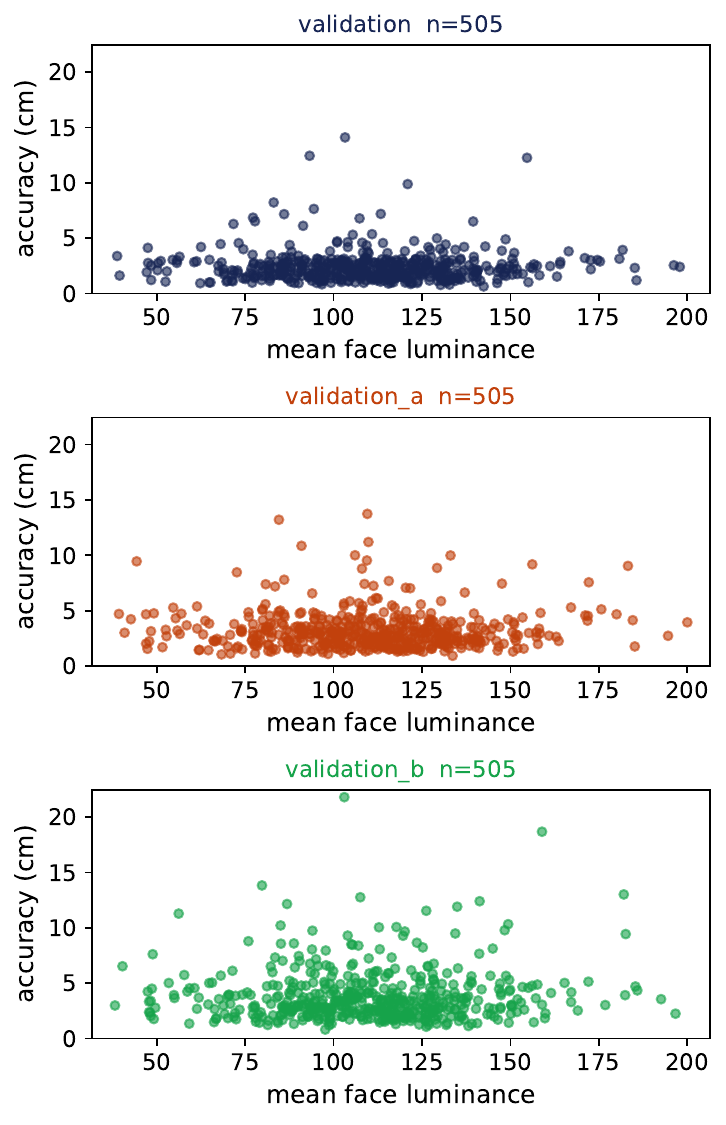}
  \caption{Chiasm mean spatial accuracy against mean face luminance on the same validation phase.}
  \label{fig:02-frame-luminance}
\end{figure}

\begin{figure}[H]
  \centering
  \includegraphics[width=\linewidth]{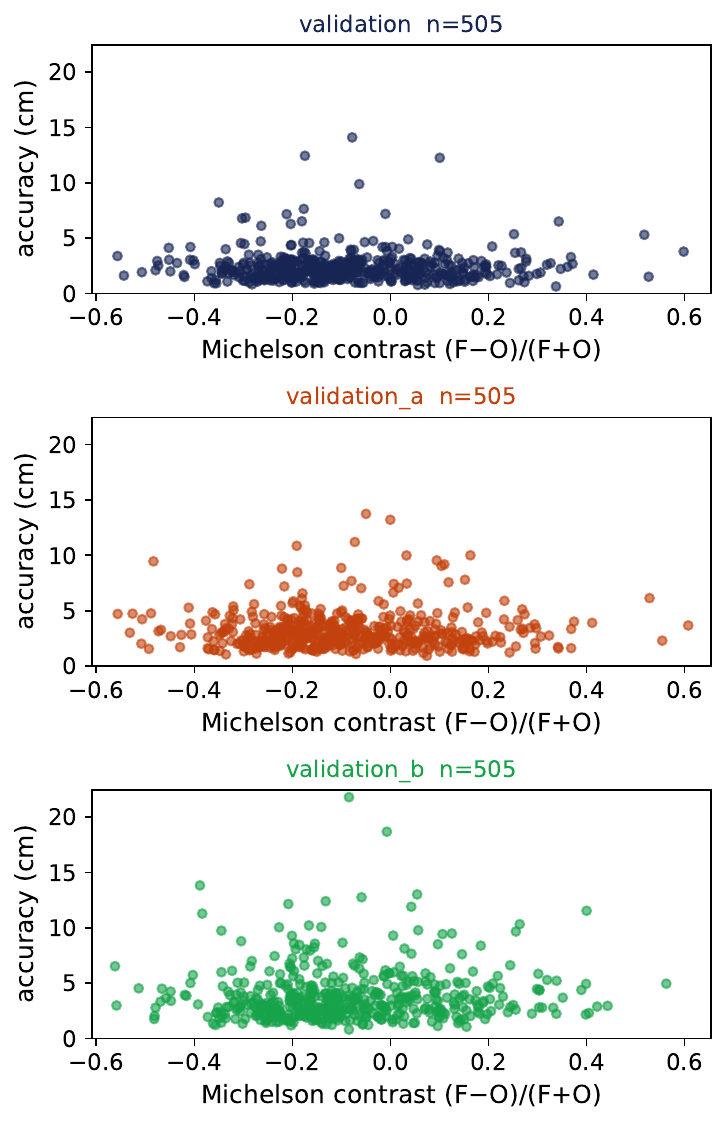}
  \caption{Chiasm mean spatial accuracy against Michelson contrast between the face and the surrounding frame.}
  \label{fig:02-frame-contrast}
\end{figure}

\section{Comparison maps}
\label{app:comparison-figures}

\begin{figure}[H]
  \centering
  \includegraphics[width=\linewidth]{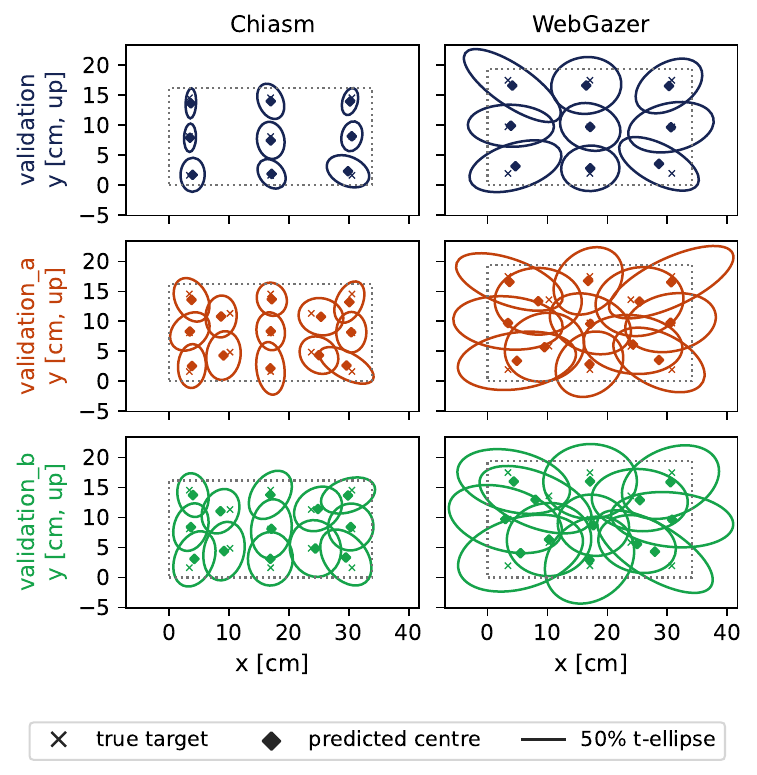}
  \caption{Spatial accuracy at each true target for Chiasm and WebGazer.
    }
  \label{fig:06-offset-map}
\end{figure}

\begin{figure}[H]
  \centering
  \includegraphics[width=\linewidth]{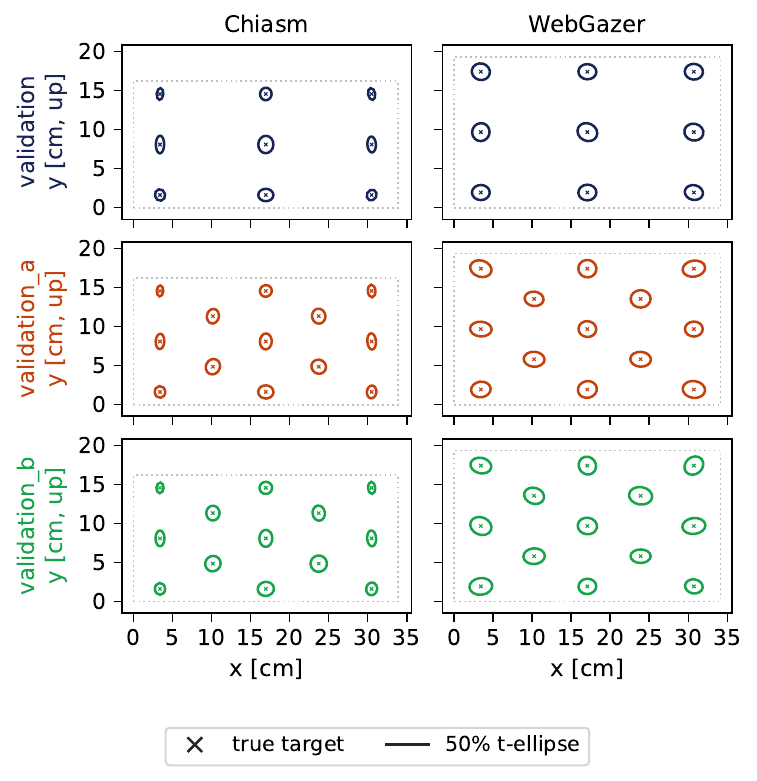}
  \caption{Within-person spatial precision (SD) at each true target for Chiasm and WebGazer.
    }
  \label{fig:06-precision-map}
\end{figure}

\begin{figure}[H]
  \centering
  \includegraphics[width=\linewidth]{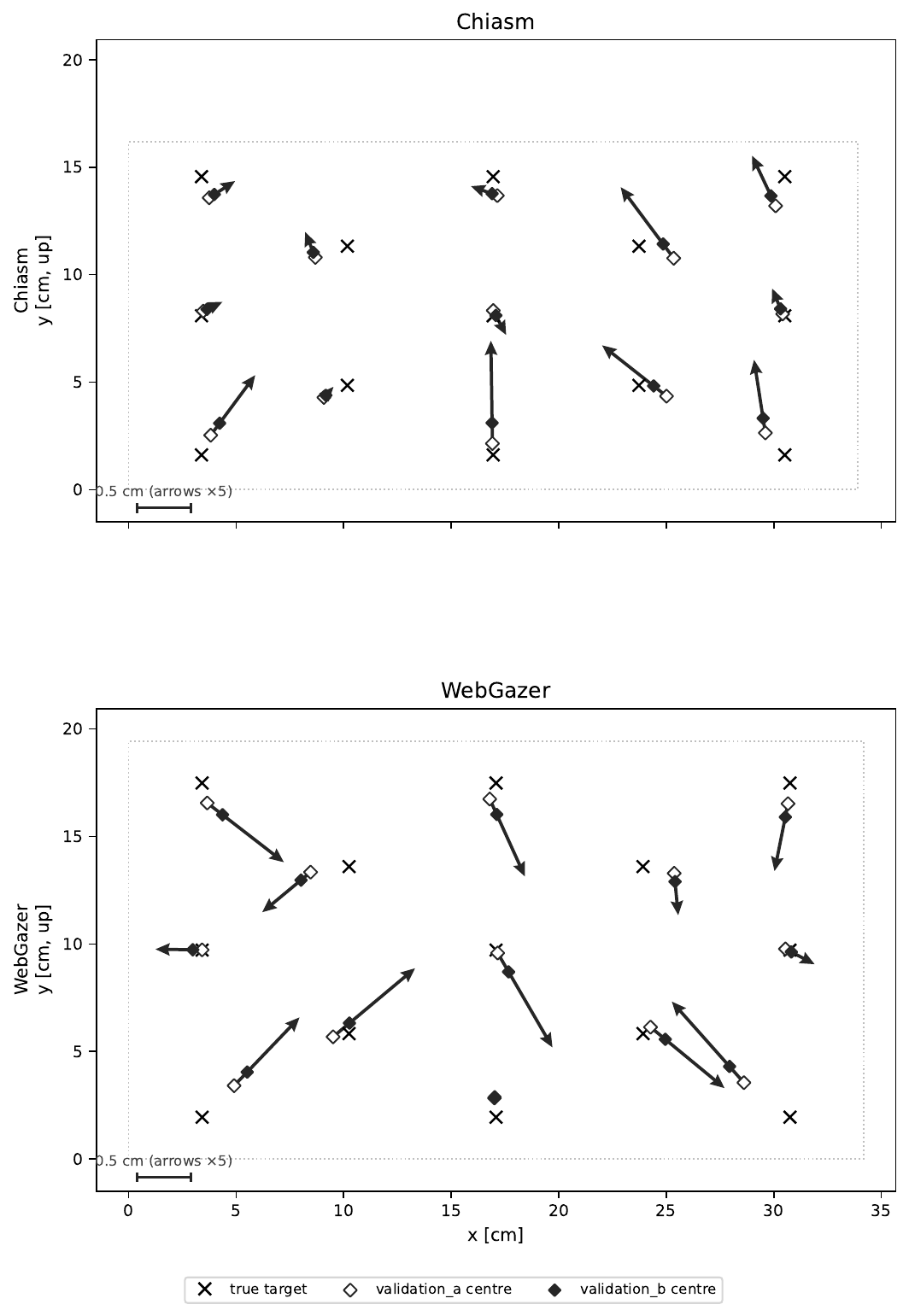}
  \caption{Drift on the dot grid, from the validation~A predicted center to the validation~B center for Chiasm and WebGazer.}
  \label{fig:06-drift-dot}
\end{figure}

\end{document}